\documentclass[fleqn,10pt]{wlscirep}
\usepackage[utf8]{inputenc}
\usepackage[T1]{fontenc}
\usepackage{lineno}
\usepackage{colortbl}  
\usepackage{xcolor}
\usepackage{array}  
\usepackage{amsmath}
\usepackage{pifont}
\usepackage{graphicx}
\usepackage{tabularx}
\usepackage{graphicx}
\usepackage{orcidlink}

\newcolumntype{Y}{>{\centering\arraybackslash}X}

\newcommand{\TQ}[1]{\textcolor{blue}{#1}}

\title{PlantSeg: A Large-Scale In-the-wild Dataset for Plant Disease Segmentation}

\author[1]{Tianqi Wei\orcidlink{0009-0005-0134-6438}}
\author[1]{Zhi Chen\orcidlink{0000-0002-9385-144X}}
\author[1]{Xin Yu\orcidlink{0000-0002-0269-5649}}
\author[2]{Scott Chapman\orcidlink{0000-0003-4732-8452}}
\author[3]{Paul Melloy\orcidlink{0000-0003-4253-7167}}
\author[1]{Zi Huang\orcidlink{0000-0002-9738-4949}}

\affil[1]{School of Electrical Engineering and Computer Science, The University of Queensland, Brisbane, 4067, Australia}
\affil[2]{School of Agriculture and Food Sustainability, The University of Queensland, Brisbane, 4067, Australia}
\affil[3]{Agriculture and Food, CSIRO, Dutton Park, Queensland, 4102, Australia}

\begin{abstract}
Plant diseases pose significant threats to agriculture. It necessitates proper diagnosis and effective treatment to safeguard crop yields. To automate the diagnosis process, image segmentation is usually adopted for precisely identifying diseased regions, thereby advancing precision agriculture. 
Developing robust image segmentation models for plant diseases demands high-quality annotations across numerous images.
However, existing plant disease datasets typically lack segmentation labels and are often confined to controlled laboratory settings, which do not adequately reflect the complexity of natural environments.
Motivated by this fact, we established PlantSeg, a large-scale segmentation dataset for plant diseases. PlantSeg distinguishes itself from existing datasets in three key aspects. (1) Annotation type: Unlike the majority of existing datasets that only contain class labels or bounding boxes, each image in PlantSeg includes detailed and high-quality segmentation masks, associated with plant types and disease names.
(2) Image source: Unlike typical datasets that contain images from laboratory settings, PlantSeg primarily comprises in-the-wild plant disease images. This choice enhances the practical applicability, as the trained models can be applied for integrated disease management.
(3) Scale: PlantSeg is extensive, featuring 11,400 images with disease segmentation masks and an additional 8,000 healthy plant images categorized by plant type. 
Extensive technical experiments validate the high quality of PlantSeg's annotations.
This dataset not only allows researchers to evaluate their image classification methods but also provides a critical foundation for developing and benchmarking advanced plant disease segmentation algorithms. 

\end{abstract}
\begin{document}

\flushbottom
\maketitle

\thispagestyle{empty}

\section*{Background \& Summary}

Plant diseases are a serious threat to agricultural productivity and can significantly impact crop yields and quality \cite{shoaib2023review}. Globally, between 20\% and 40\% of all crops are lost due to plant diseases. According to The Food and Agriculture Organization of the United Nations \cite{plant_pathology}, annual losses exceed 220 billion dollars due to plant diseases. 
Early and accurate plant disease detection and assessment is crucial for minimizing economic losses.
Traditionally, manual diagnosis by plant pathologists is considered the most reliable method of assessment. However, diagnosticians are not always available to provide assessment in a timely manner, leading to potentially costly delays. 
Further, plant pathologists are often skilled at recognizing a limited number of plant diseases on a handful of hosts, thus multiple plant pathologists or taxonomists are usually required for a reliable diagnosis. 

Arguably, one of the goals for precision agriculture \cite{shafi2019precision} includes improvements to agricultural systems enabling the automatic localization and segmentation of disease-affected plants and plant parts. 
Generic image segmentation methods \cite{chen2017deeplabv3, chen2018deeplabv3+, kirillov2020pointrend, guo2022segnext} have demonstrated outstanding performance on commonly used benchmark datasets, such as ADE20k \cite{ade20k}, Cityscapes \cite{cordts2016cityscapes} and MSCOCO \cite{coco}.
However, there is still a huge gap between the mainstream segmentation models and the common ones being used for plant disease segmentation. 
Most recent plant disease segmentation studies \cite{bhatti2024advanced, jafar2024revolutionizing, wang2021tcnn, li2022improvedMSMSVDD, xie2020deep, savarimuthu2021investigation, snap, shoaib2022deep} typically adopt obsolete deep learning models for segmenting narrow selections of host and pathogen relationships. 
In contrast, a more generalized approach to segmenting a wider variety of plant diseases sets a far more fine-grained and challenging task, to model the characteristics of different diseases.

The challenge of developing an advanced deep learning-based plant disease segmentation model is made more difficult due to the substantial number of annotated plant images required and the lack of publicly available high-quality datasets. Currently, the availability of plant disease datasets is limited, and most accessible datasets are not sufficiently labeled in terms of annotation type, image source, and scale. We provided the statistics of existing plant disease datasets in Table \ref{tab:my_label} and elaborate on their insufficiency as follows:

\begin{figure*}[t]
    \includegraphics[width=0.9\textwidth]{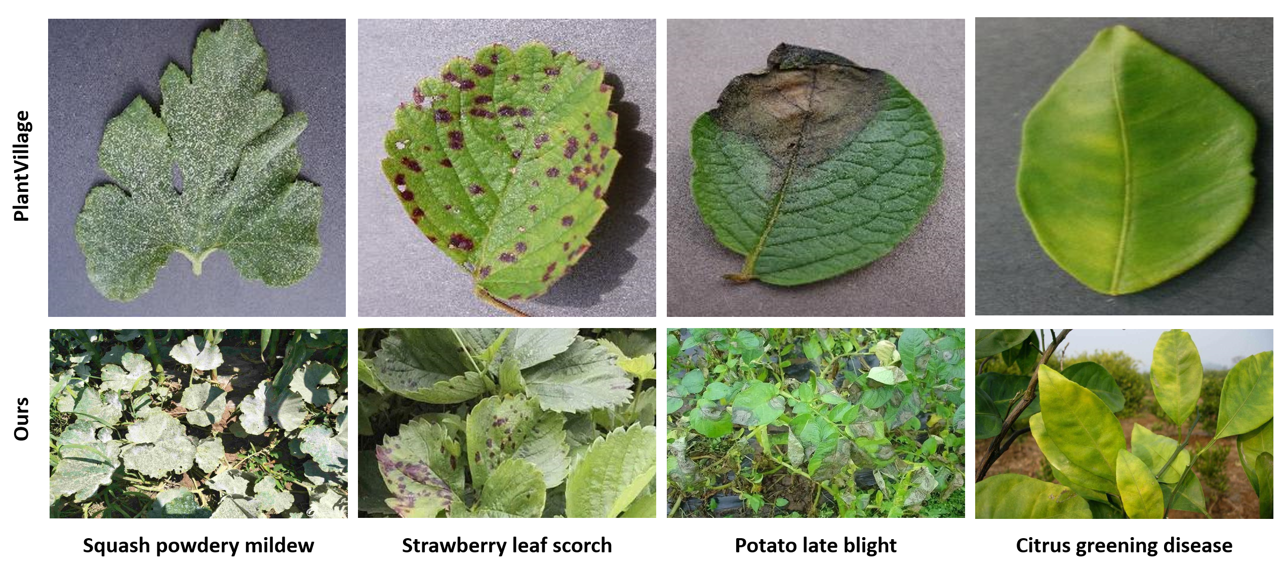}
    \caption{Examples of images of PlantVillage \cite{plantvillage} and our dataset. As collected in laboratory environments, each image in PlantVillage only contains one leaf and has a uniform background, while images of our dataset feature much more complex backgrounds, various viewpoints, and different lighting conditions.}
    \label{fig:labimage}
\end{figure*}

\begin{table}[t]
\resizebox{0.9\linewidth}{!}{
    \renewcommand\arraystretch{1.5}{
    \begin{tabular}{|l|l|l|l|l|c|c|c|l|}
    \hline
     \rowcolor{orange!50}    \textbf{Dataset Name} & \textbf{Year} & \textbf{\#Images} & \textbf{\#Classes} & \textbf{\#Plants} & \textbf{In-the-wild} & \textbf{Bounding box} & \textbf{Segmentation mask}& \textbf{References}\\
    \hline
          PlantVillage & 2015 & 54,309 & 38 & 14 & \ding{55} & \ding{55} & \ding{55} & Hughes, \textit{et al.} \cite{plantvillage}\\
    \hline
         PlantDoc & 2020 & 2,598 & 27 & 13 & \ding{51} & \ding{55} & \ding{55} & Singh,  \textit{et al.}\cite{singh2020plantdoc}\\
    \hline
    FieldPlant & 2023 & 5,170 & 27 & 3 & \ding{51} & \ding{55} & \ding{55} & Moupojou,  \textit{et al.}\cite{moupojou2023fieldplant}\\
    \hline
     PlantWild & 2024 & 18,542 & 89 & 33 & \ding{51} & \ding{55} & \ding{55} & Wei,  \textit{et al.} \cite{MVPDR}\\
    \hline
         Tomato Disease & 2020 & 4,178 & 4 & 1 & \ding{55} & \ding{51} & \ding{55} & Zhang, \textit{et al.} \cite{zhang2020tomato_dataset}\\
    \hline
        
        GLDD &  2020 & 4,449 & 4 & 1 & \ding{55} & \ding{51}& \ding{55} & Xie, \textit{et al.} \cite{xie2020deep}\\
    \hline
    APPLE \& GRAPE& 2021 & 1,150 & 6 & 2 & \ding{55} & \ding{51} & \ding{55} & Savarimuthu,\textit{et al.} \cite{savarimuthu2021investigation}\\
    \hline
    MSMSVDD & 2022 & 1,000 & 5 & 3 & \ding{51} & \ding{51} & \ding{55} & Li, \textit{et al.}\cite{li2022improvedMSMSVDD}\\
    \hline
     LDSD & 2021 & 588 & 1 & N/A & \ding{51} & \ding{55} & \ding{51} & Fakhre, \textit{et al.}\cite{kaggleLeafDisease}\\
    \hline
     NLB & 2024 & 1,000 & 1 & 1 & \ding{51} & \ding{55} & \ding{51} & Prashanth, \textit{et al.}\cite{prashanth2024towards}\\
    \hline

   \textbf{PlantSeg (ours)} & 2024 & 11,458 & \textbf{115} & \textbf{34} & \ding{51} 
 & \ding{55} & \ding{51} & N/A\\
    \hline
    \end{tabular}}}
    \caption{Summary of plant disease image datasets. Existing datasets }
    \label{tab:my_label}
\end{table}

\begin{itemize}
\item{\textbf{Annotation Type.} Most existing plant disease image datasets are designed for classification or object detection tasks. Classification involves identifying the global content in an image but does not provide any local information. Object detection can localize objects by drawing bounding boxes around them. There are only a few datasets available for image segmentation \cite{kaggleLeafDisease,prashanth2024towards}, which still require detailed and precise mask annotations. Accurately localizing the affected areas, pixel-wise, with segmentation masks provides improved granularity compared to purely predicting the bounding box with object detection \cite{zhang2020tomato_dataset,xie2020deep,savarimuthu2021investigation,li2022improvedMSMSVDD}. The affected areas are usually irregular shapes, the outputs from semantic segmentation can be directly used in automated systems for precision agriculture, such as quarantining affected areas within paddocks, variable fungicide application rates to minimize the spread through the paddock. These assessments might lead to the ability to use decision support tools and integrated disease management (IDM) on a sub-paddock scale.
}
\item{\textbf{Image Source.} Images in many existing datasets \cite{plantvillage, zhang2020tomato_dataset, xie2020deep, savarimuthu2021investigation} are collected under controlled laboratory conditions. As illustrated in the upper row of Figure \ref{fig:labimage}, such images lack the complex background information in field environments. As a result, algorithms trained on `staged' datasets may not perform in real-world scenarios where plants are subjected to various environmental factors, such as varying lighting conditions, occlusions, and background noise. This makes lab-trained models unsuitable for segmentation tasks in the field where they are likely to be applied for integrated disease management.}
\item{\textbf{Scale.} Existing datasets are often small in scale \cite{kaggleLeafDisease, prashanth2024towards}, which contain a limited number of images, categories, and a narrow host and pathogen focus. Consequently, algorithms trained on such datasets cannot be applied to detect diseases outside the scope for which they have been characterized as their lack of generalizability limits the practical application.}
\end{itemize}

To address these problems in plant disease segmentation research, we present PlantSeg, the largest dataset for plant disease segmentation in the wild. It has the most number of categories among all existing plant disease datasets. PlantSeg contains 11,458 images of 115 disease categories with corresponding segmentation annotations. The segmentation annotation is carried out by trained annotators and checked by expert pathologists to ensure accuracy. This paper showcases the characteristics of PlantSeg and benchmarks state-of-the-art segmentation models on plant disease segmentation. We believe our dataset can serve as a comprehensive benchmark for developing plant disease segmentation methods.

\section*{Methods}
\subsection*{Image acquisition}
\begin{figure*}[t]
    \resizebox{0.9\width}{0.8\height}{
    \includegraphics[width=\textwidth]{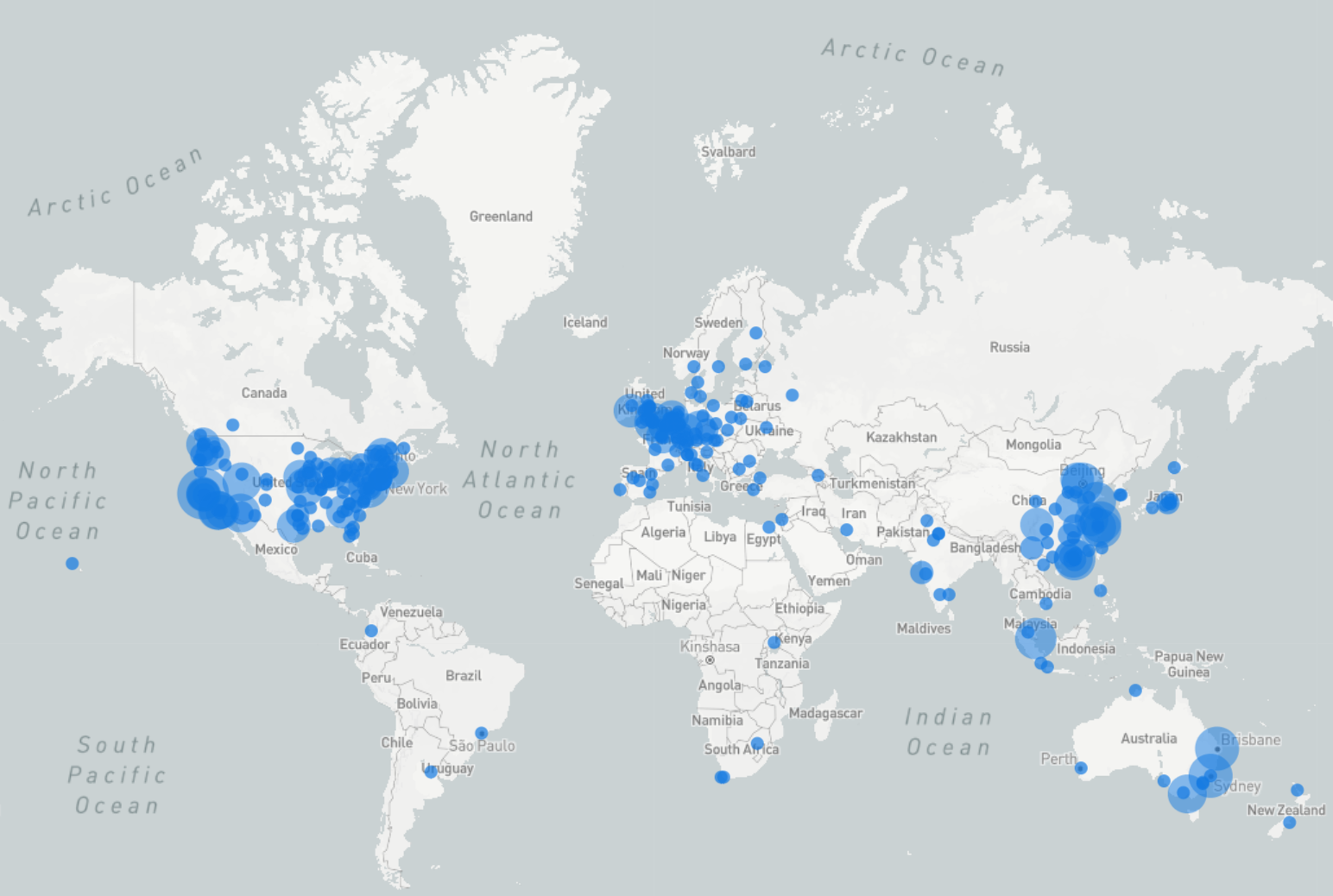}}
    \caption{Locations of the source image acquired. The sizes of the plots represent the number of acquired images. The size of each circle demonstrates the number of images acquired from the address, and the color depth indicates the density of addresses within a nearby region.}
    \label{fig:ip}
\end{figure*}

To build our dataset, we carefully selected plants that hold substantial economic and nutritional significance. It consists of \textbf{profit crops} with high commercial value, \textbf{staple crops} that are essential for human consumption, as well as a diverse range of \textbf{fruits} and \textbf{vegetables}, which contribute significantly to agricultural production. 
Including a wide variety of plants, our dataset is both comprehensive and representative of the most important plants in the agriculture field. Consequently, we have identified 115 diseases across 34 plants for our dataset curation. 
For each plant disease, we utilized its name as the keyword to search relevant images from various internet sources, including Google Images, Bing Images, and Baidu Images. This comprehensive collection strategy can broaden the retrieval range and increase the diversity of results, as images were collected from websites all over the world. Therefore, the scale of our dataset was effectively expanded.
The IP addresses of the image source websites are presented in Figure \ref{fig:ip}, and show our dataset incorporates images from diverse regions from around thee world.

\subsection*{Image cleaning}
The retrieved images were organized into folders corresponding to their respective disease names. Data quality was 'cleaned' by a review process before proceeding with segmentation annotation. The cleaning process involved our annotators carefully reviewing each image and removing any incorrect images in a class folder, retaining only accurate images of each class. 
Ambiguous images, or images difficult to classify, were also discarded. 
This process ensured the accuracy of every image through cross-validation by at least two annotators. Furthermore, in cases where there were discrepancies between the annotators' judgements, experts with extensive knowledge and experience were brought in to review and make final decisions.

\subsection*{Segmentation annotation}

\begin{figure*}[t]
    \includegraphics[width=0.9\textwidth]{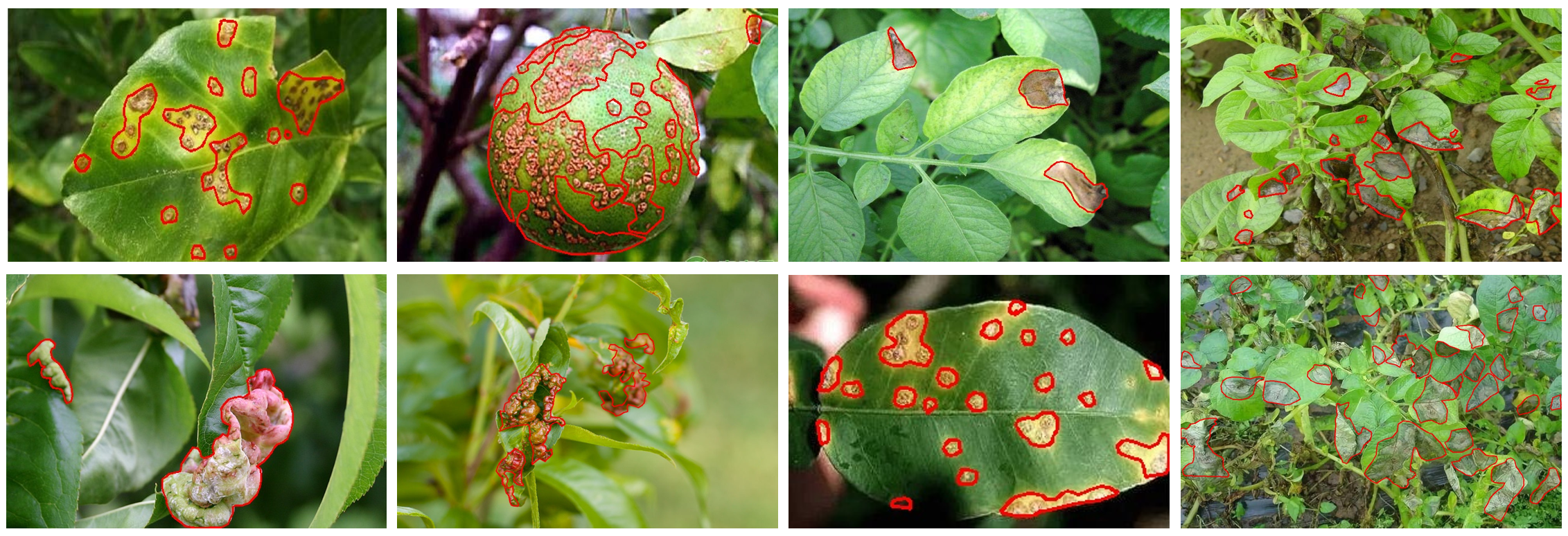}
    \caption{Examples of images with annotated polygons on the disease-affected areas.}
    \label{fig:annotation_examples}
\end{figure*}

After the review process, we established a segmentation annotation standard to ensure consistent labelling of disease-affected areas in the images. For distinct and sizeable lesions, annotations were made with individual polygons. Overlapping lesions were annotated as a combined affected area. For diseases like rust and powdery mildew, which present as small, densely clustered symptoms on both leaves and fruits, we meticulously annotated the infected regions to accurately reflect the disease distribution. Additionally, any deformities in plant leaves or fruits caused by diseases were also annotated. Examples of annotated images are shown in Figure \ref{fig:annotation_examples}.

Under the guidance of two expert plant pathologists, a group of annotators participated in the segmentation annotation process. The annotators were trained on the annotation standard and then required to annotate 10 images for evaluation. The expert pathologists reviewed these annotations, and any annotator whose work was deemed unsatisfactory was asked to re-annotate the images. Annotators who consistently did not meet the re-annotation standards were disqualified from further annotation tasks. Conversely, if the annotation results met the standard, the annotators were approved to proceed to the subsequent stages of the project. 

\begin{figure*}[t]
    \includegraphics[width=0.95\textwidth]{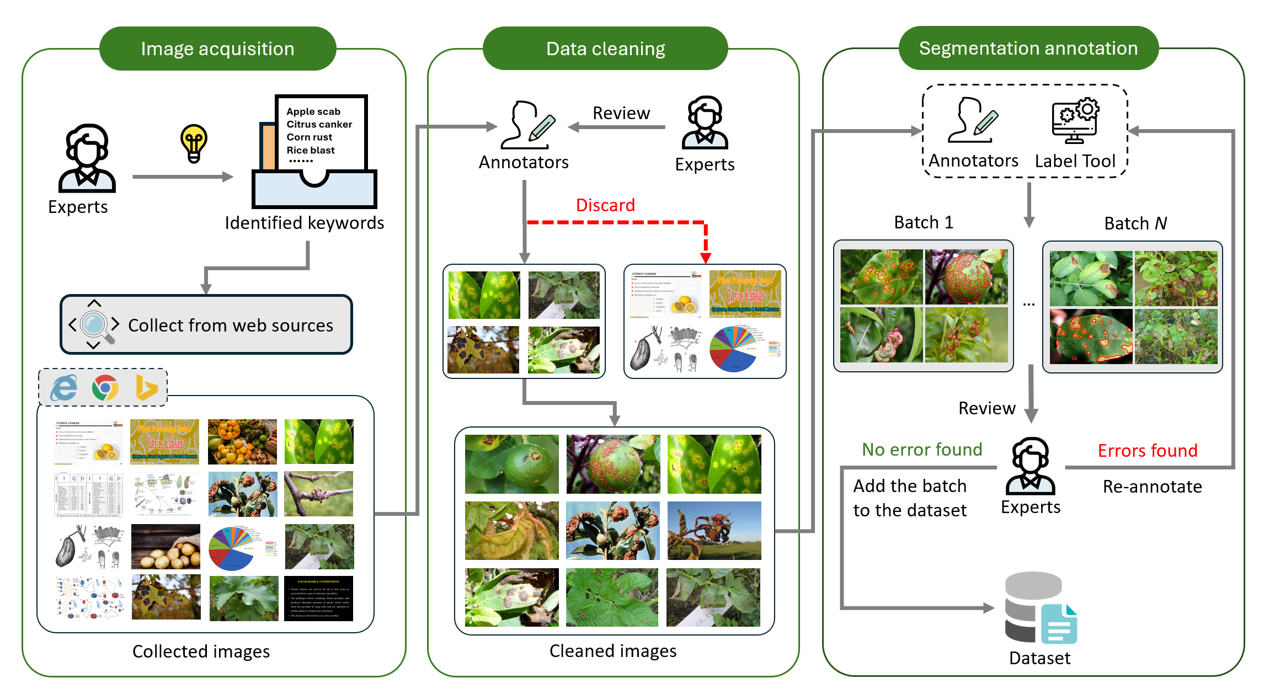}
    \caption{The curation process of the PlantSeg dataset involves three main steps: image acquisition, data cleaning, and annotation. In the image acquisition stage, images were collected from various internet sources using identified keywords and then stored according to their categories. During the data cleaning phase, incorrect images were identified and removed. For the segmentation annotation process, annotators utilized LabelMe \cite{labelme} to annotate the cleaned images. These annotations were subsequently reviewed by experts and saved in JSON files.}
    \label{fig:workflow}
\end{figure*}

In total, 10 annotators were engaged in this detailed segmentation annotation work. We divide the images into various subsets. 
Images were subsetted by host plant and included a number of different disease.
Each annotator was assigned a specific subset and used LabelMe (V5.5.0) \cite{labelme} software to annotate the diseased parts. 
Following the image subset primary annotations, by the first annotator, the image subset was passed to another annotator for review, and correct any errors. 
The final annotations underwent a rigorous review by expert pathologists.
Figure \ref{fig:workflow} demonstrates the whole workflow of the segmentation annotation process.


\renewcommand\arraystretch{1.5}{
\begin{table}[t]
    \begin{tabular}{|l|l|}
    \hline
        \rowcolor{orange!50} \textbf{Metadata} & \textbf{Description}\\
        \hline
         Name & The name of the image\\
         \hline
         Plants & The host plants of interest\\
         \hline
         Diseases & Disease type\\
         \hline
         Resolutions & The resolution of the image\\
         \hline
         Label files & The path of corresponding label file\\
         \hline
         Mask ratios & The proportion of annotated pixels to the total number of pixels.\\
         \hline
         URLs & Download link of the image if available\\
         \hline
         Training/Test split \qquad \qquad & Specify images of each disease type for Training/Test set \qquad \qquad\\
         \hline
    \end{tabular}
    \caption{The Metadata of PlantSeg.}
    \label{tab:metadata}
\end{table}}

\subsection*{PlantSeg metadata}
In this section, we introduce the metadata of the PlantSeg dataset, which provides detailed information related to each image. An overview is presented in Table \ref{tab:metadata}.
\\

\noindent \textit{\textbf{Plants and Diseases.}}  \hspace{5pt}
34 host plants of interest were determined according to the suggestions of experts. 
These were classified into three categories relating to their socioeconomic importance to humans.
Profit crops, included commodities with high market demand and low nutritional value, \textit{e.g.,} Coffee and Tobacco. Staple crops, refer to crops providing the majority of people's carbohydrate and protein requirements, such as wheat, corn, and potatoes.  PlantSeg also contains a wide range of fruits and vegetables, such as apples, oranges, and tomatoes, which supplement nutritional needs and are vital for people's diets. Based on the plants of interest, 115 common disease categories were determined to be included in the dataset.
\\

\noindent \textit{\textbf{URLs.}} \hspace{5pt}
To aid in the reproducibility of this work, all downloaded images used to build the PlantSeg dataset were stored with URL links to the source websites. This allows researchers to verify the source of the images and ensure compliance with copyright regulations.
\\

\noindent \textit{\textbf{Label files.}} \hspace{5pt}
The label files share the same filenames as their corresponding images, differing only by their file extensions. These labels are stored as grayscale PNG images, where pixels representing diseased areas are annotated with specific class index values, while all other pixels are assigned a value of zero.
\\

\noindent \textit{\textbf{Image resolutions and Mask ratios.}} \hspace{5pt}
The image resolution indicates the image size including width and height, and the mask ratio denotes the proportion of labeled pixels to the total number of pixels in the image.
\\

\noindent \textit{\textbf{Training/Test sets of PlantSeg.}} \hspace{5pt}
PlantSeg is built to evaluate segmentation methods on plant disease images. We randomly selected 20\% of the images from each disease as the test set, while the remaining images were used as the training set.



\section*{Data Records}
The PlantSeg dataset can be downloaded through \href{https://doi.org/10.5281/zenodo.13293891}{\TQ{https://doi.org/10.5281/zenodo.13293891}} \cite{zenodo}. The repository is covered under the CC BY-NC-ND 4.0 licence. Plant disease images are saved in JPEG format and are stored in the ``images'' folder, and the labels are saved in PNG format and are stored in the ``annotations'' folder. Each image and its corresponding label have the same file name except the file extension. In addition, the original label files generated by LabelMe \cite{labelme} and saved in JSON format, are provided in the ``json'' folder. All images and labels are split into training and test sets with an 80/20 ratio. A PlantSeg-Meta.csv file is provided to store the meta-information presented in Table \ref{tab:metadata}.


\section*{Technical Validation}

\subsection*{Data property analysis}
We conduct a comprehensive analysis of the PlantSeg dataset from multiple aspects, including distributions of disease type, image resolution, and segmentation mask ratio.
\\

\begin{figure*}[t]

    \includegraphics[width=0.6\textwidth]{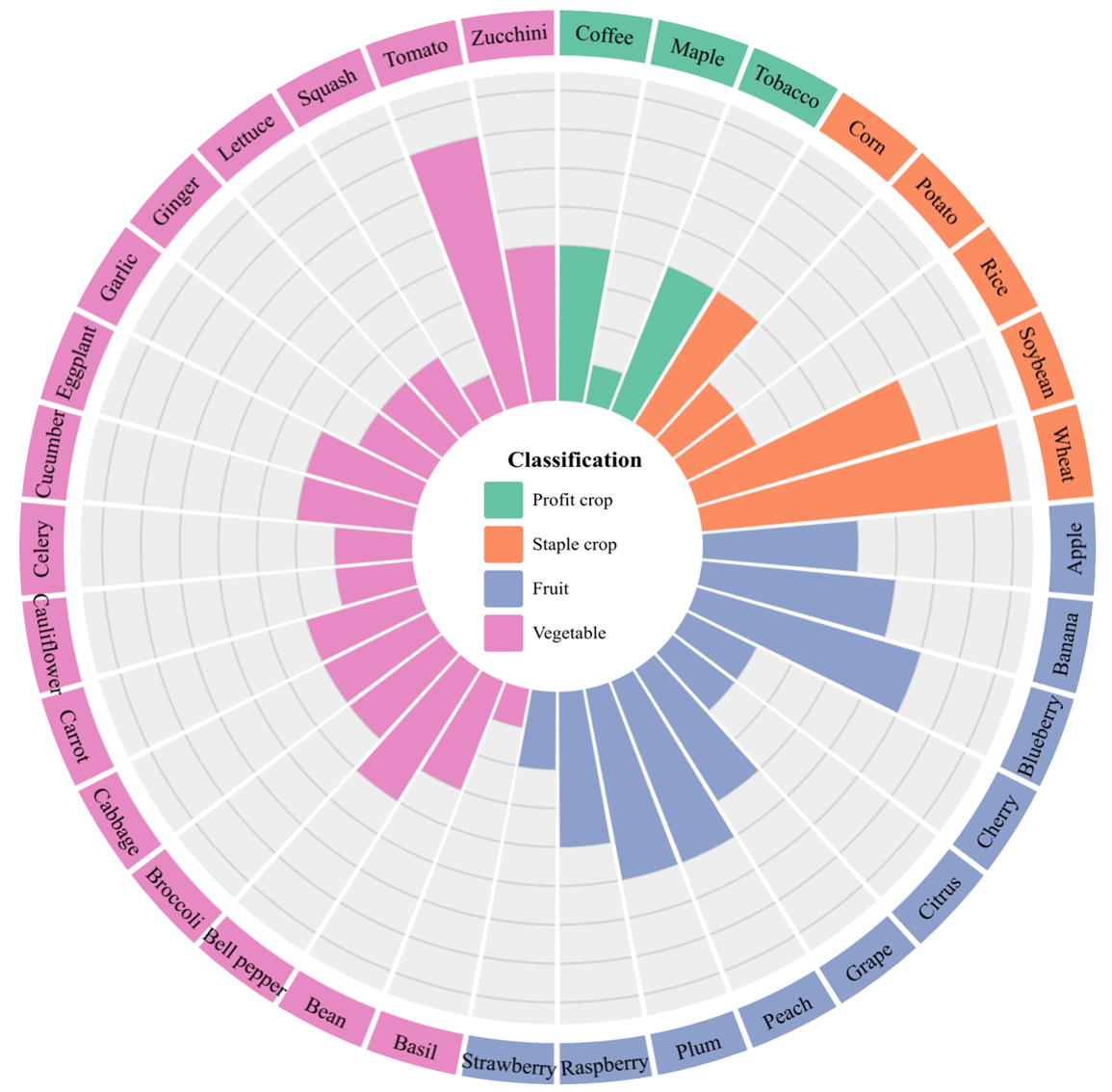}
    \vspace{3pt}
    \caption{Disease distribution in PlantSeg according to plants and Socioeconomic classification. The height of the bars represents the number of diseases associated with each plant.}
    \label{fig:disease_type_distribution}
\end{figure*}

\begin{figure*}[t]
    
    \includegraphics[width=0.9\textwidth]{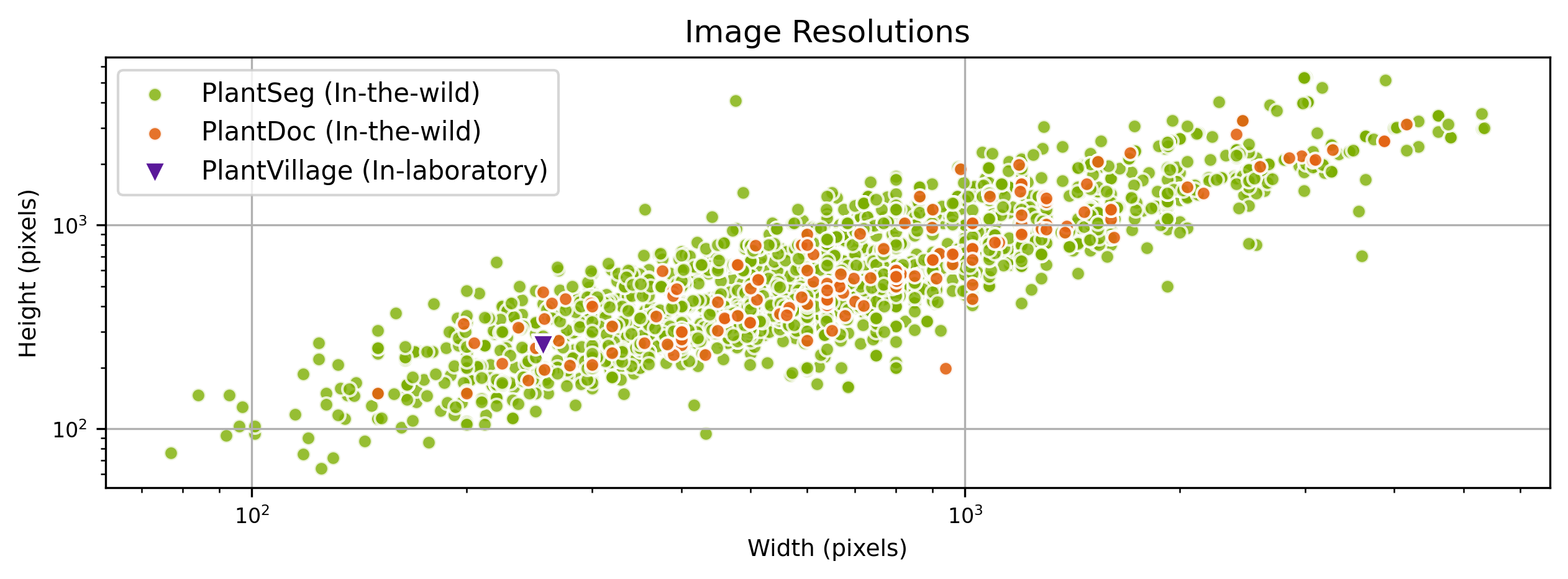}
    \vspace{-1.2em}
    \caption{The resolution distributions of different datasets, including PlantVillage, PlantDoc and PlantSeg. Each green dot represents a single image in the PlantSeg database; Red, for each image in the PlantDoc database and inverted yellow triangles for the PlantVillage database.}
    \label{fig:distribution}
\end{figure*}
\noindent \textit{\textbf{Plant and disease type distribution.}} \hspace{5pt}
The PlantSeg dataset includes 115 diseases across 34 plant hosts, which are categorized into four major socioeconomic groups: \textbf{profit crops}, \textbf{staple crops}, \textbf{fruits}, and \textbf{vegetables}. Figure \ref{fig:disease_type_distribution} provides an overview of the distribution of plant hosts and disease types. Fifteen plant hosts from vegetables and ten fruit hosts constitute significant portions of the dataset, with 45 and 39 diseases respectively. In contrast, profit crops consist of only 9 diseases across 3 plant hosts, accounting for 7.8\% of the total image database.
\\


\noindent \textit{\textbf{Image resolution distribution.}} \hspace{5pt}
We analyze image resolution distribution in PlantSeg and compare it with two widely used plant disease image datasets: PlantVillage\cite{plantvillage} and PlantDoc \cite{singh2020plantdoc}. PlantVillage consists of more than 50,000 images, all images are captured from plant hosts under controlled experimental conditions. PlantDoc contains images collected from field environments, and only includes about 2,600 images. The resolution distribution is presented in Figure \ref{fig:distribution}. This scatter plot demonstrates that PlantSeg covers a wide range of image resolutions and reflects the variability typical of real-world conditions, compared to other databasses. PlantDoc (red points) also exhibits considerable variability, albeit on a lower scale compared with PlantSeg. 
PlantVillage images were curated from uniform laboratory settings with a single resolution. All data points overlap in a single point on the plot.
Figure \ref{fig:distribution} reveals the diversity and range of image resolutions in in-the-wild datasets compared to laboratory-collected data. It indicates the challenge when working on real-world data collection and the inherent variance in images. In addition, it also emphasizes the advantage of a dataset, such as ours, which focuses on field-based images with variable scale and diversity.
\\

\begin{figure*}[tb]

    \includegraphics[width=0.9\textwidth]{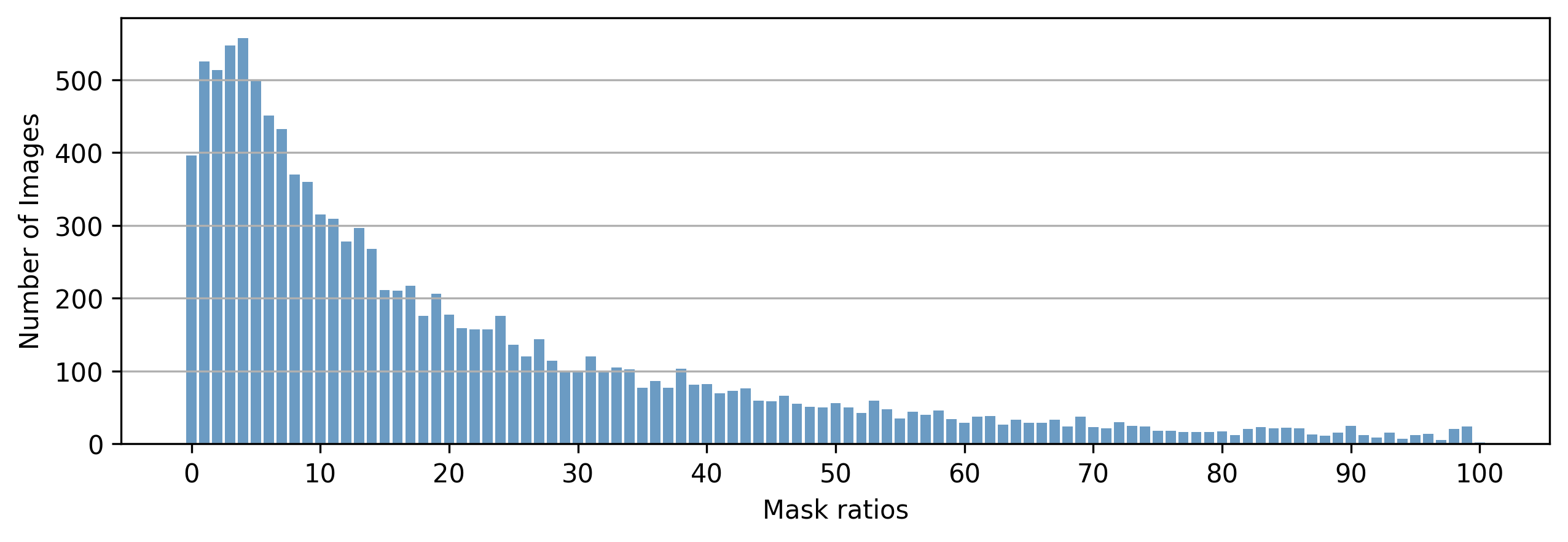}
    \caption{The horizontal axis represents the percentage of mask area relative to the entire image, while the vertical axis represents the number of corresponding images.}
    \label{fig:mask_distribution}
\end{figure*}

\begin{figure*}[tb]
    \includegraphics[width=0.9\textwidth]{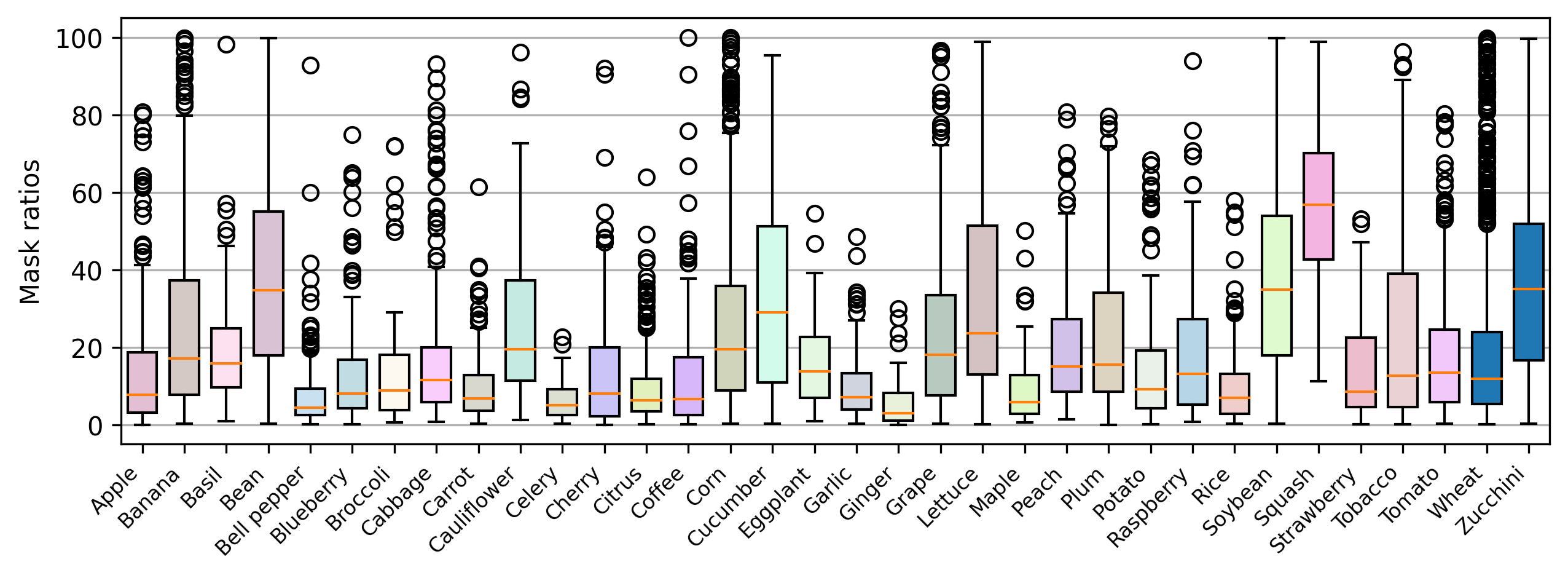}
    \caption{The boxplot shows the percentage distribution of mask area per image, which varies significantly among different plants.}
    \label{fig:box_plot}
\end{figure*}

\noindent \textit{\textbf{Segmentation mask ratio distribution.}} \hspace{5pt}
Figure \ref{fig:mask_distribution} depicts the distribution of segmentation mask ratios in the PlantSeg dataset. A low ratio indicates the annotated disease area is relatively small, while a high ratio suggests a larger area. Overall the mask ratio of the PlantSeg image database exhibits significant variation. The box plot in Figure \ref{fig:box_plot} shows considerable variation in mask ratio distribution among different diseases and within each disease type.

\subsection*{Evaluation on PlantSeg}
To establish benchmarks for plant disease segmentation, we applied four segmentation methods and evaluated their performance on the PlantSeg dataset.
\\

\noindent \textit{\textbf{Baseline models.}} \hspace{5pt}
We employed four commonly used and state-of-the-art semantic segmentation methods as our baselines, including Side Adapter Network (SAN) \cite{xu2023san}, DeepLabv3 \cite{chen2017deeplabv3}, DeepLabv3+ \cite{chen2018deeplabv3+}, and SegNeXt \cite{guo2022segnext}.
 These methods were trained and evaluated on the same training and test sets respectively. They were implemented using PyTorch version 1.11. For DeepLabv3 \cite{chen2017deeplabv3},  DeepLabv3+ \cite{chen2018deeplabv3+}, and SAN \cite{xu2023san}, we respectively leveraged different variants of ResNet \cite{resnet} and Vision Transformer \cite{vit} as the backbones. All methods were trained using a Stochastic gradient descent (SGD) optimizer, with a learning rate of 0.001, a momentum of 0.9, and a weight decay rate of 0.0005. We introduced cross-entropy loss to optimize the models. Each model was trained with a fixed batch size of 16.
\\

\noindent \textit{\textbf{Evaluation metrics.}} \hspace{5pt}
 We introduced the Mean Intersection over Union (MIoU) and Mean Accuracy (mAcc) as our evaluation metrics. MIoU calculates the average Intersection over Union across all classes, offering insight into the model's overall segmentation performance. Mean Accuracy measures the proportion of correctly classified pixels within each class and then averages the accuracies across all classes. These evaluation metrics are computed as:
\begin{equation}
\text{MIoU}_i =\frac{1}{N} \sum_{i=1}^{N} \frac{TP_i}{TP_i + FP_i + FN_i}, \qquad
\text{mAcc} = \frac{1}{N} \sum_{i=1}^{N}\frac{TP_i}{TP_i + FN_i},
\end{equation}
where $N$ represents the number of classes. $TP_i$, $FP_i$, $TN_i$, $FN_i$ are the number of true positive, false positive, false negative, and true negative pixels of the $i$-th class respectively. 
\\


\noindent \textit{\textbf{Experiment results.}} \hspace{5pt}
The segmentation results of the selected baselines are summarized in Table \ref{tab:comparison}. The findings show that DeepLabv3 \cite{chen2017deeplabv3} and DeepLabv3+ \cite{chen2018deeplabv3+} with a deeper ResNet-101 backbone outperform those with a ResNet-50 backbone in both MIoU and mAcc scores. Similarly, SAN \cite{xu2023san} with a ViT-L/14 backbone demonstrates superior performance compared to the ViT-B/16 backbone, with a 4.79\% increase in MIoU and a 5.89\% increase in mAcc.  This suggests that using a larger backbone can benefit the methods and lead to substantial improvements in segmentation performance.
Among all the baselines, SegNeXt \cite{guo2022segnext}, which incorporates a large multi-scale convolutional backbone, achieves the highest performance, with the MIoU of 53.89\% and the mAcc of 65.91\%.

\begin{table}[tb]
    \tabcolsep=0.5cm
    \renewcommand\arraystretch{1.5}{
    \begin{tabularx}{0.9\textwidth}{|Y|Y|Y|Y|}
        \hline
        \rowcolor{orange!50} \textbf{Method} & \textbf{Backbone} & \textbf{MIoU}  & \textbf{mAcc}\\
        \hline
        DeepLabv3 \cite{chen2017deeplabv3} & ResNet-50 \cite{resnet} & 17.24  & 37.95\\
        \hline
        DeepLabv3 \cite{chen2017deeplabv3} & ResNet-101 \cite{resnet} & 20.72  & 40.63\\
        \hline
        DeepLabv3+ \cite{chen2018deeplabv3+}& ResNet-50 \cite{resnet} & 25.08  & 40.66\\
        \hline
        DeepLabv3+ \cite{chen2018deeplabv3+}& ResNet-101 \cite{resnet} & 27.18  & 42.29\\
        \hline
        SAN \cite{xu2023san} & ViT-B/16 \cite{vit} & 34.79  & 50.19\\
        \hline
        SAN \cite{xu2023san} & ViT-L/14 \cite{vit} & 36.91  & 52.81\\
        \hline
        SegNeXt \cite{guo2022segnext} & MSCAN-L & 44.52  & 59.95\\
        \hline
        
    \end{tabularx}}
    \caption{Performance comparison of different methods on PlantSeg.}
    \label{tab:comparison}
\end{table}

\begin{figure*}[tb]
    \includegraphics[width=0.9\textwidth]{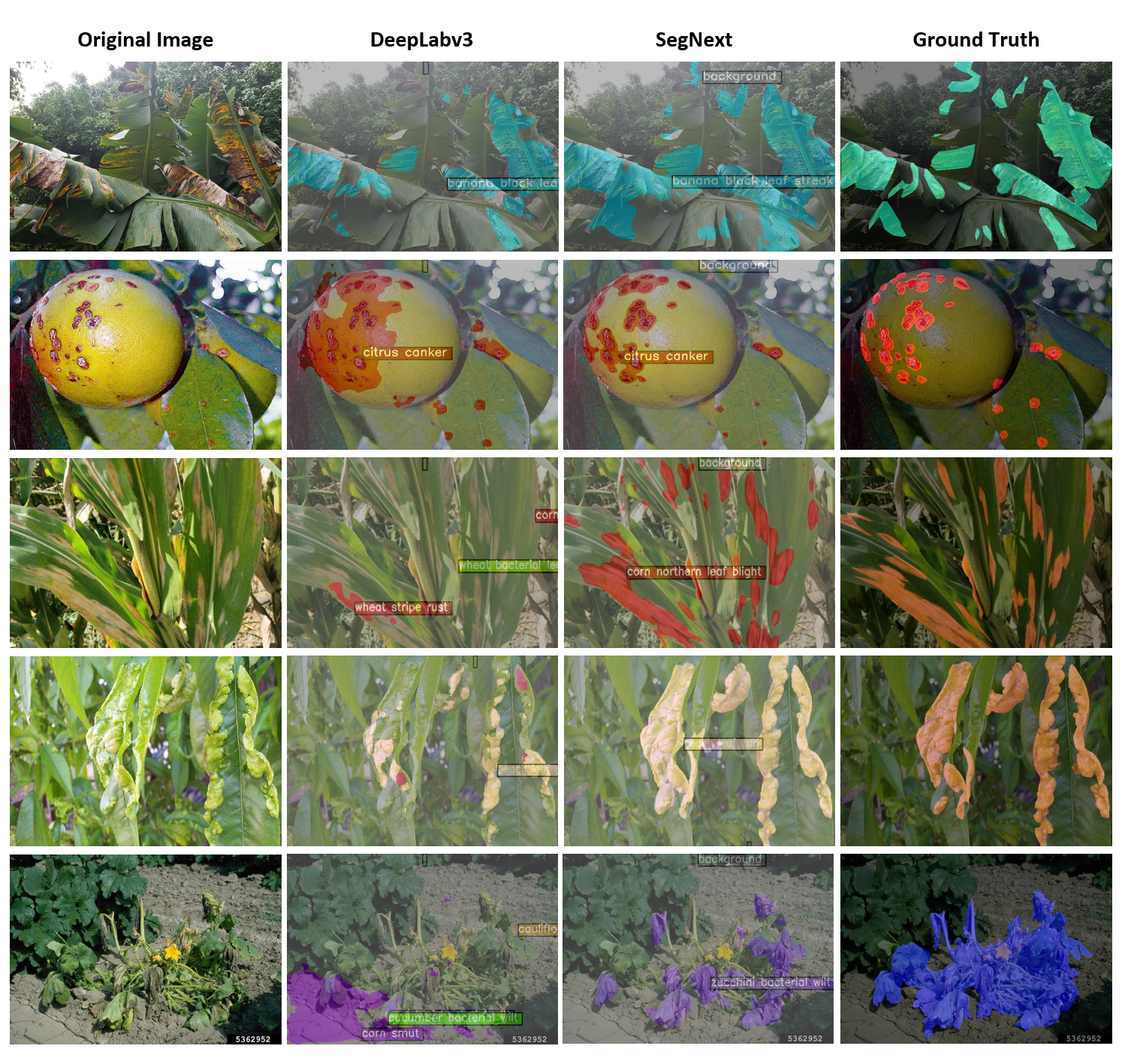}
    \caption{Visualization of some experimental results on the test set of PlantSeg. From left to right: image examples of PlantSeg, results of DeepLabv3 \cite{chen2017deeplabv3}, results of SegNext \cite{guo2022segnext}, and the ground truth annotations.}
    \label{fig:seg}
\end{figure*}

Figure \ref{fig:seg} presents a series of visualizations comparing the ground truth masks with the predictions from various baselines, including DeepLabv3 \cite{chen2017deeplabv3} and SegNext \cite{guo2022segnext}. The results from DeepLabv3 are particularly unsatisfactory, as it struggles to identify and locate the diseased areas accurately. The state-of-the-art SegNext delivers more accurate results than DeepLabv3, as it can effectively segment lesions and deformation on leaves and fruit according to the results shown in the 1st to 4th rows of Figure \ref{fig:seg}. 
However, in the cases of the 5th row, SegNeXt focuses on the wilted leaves but overlooks the collapsed stems. This suggests that segmentation becomes more challenging when the disease involves curling and deformation of stems.

\subsection*{Conclusion}
PlantSeg offers a step forward towards automation for disease detection and quantification using ordinary proximal sensing devices such as RGB cameras.
Multispectral and hyperspectral devices, which are currently used in research for disease detection and quantification in the field have a higher cost and are beyond the budgets for many in the agricultural industry \cite{bock2020visual}. 
Segmentation of images to define the area of affected plant parts can allow an automated method for quantification of diseases signs and symptoms at a stated point in time. 
This provides an unbiased method for estimating disease severity scores for researchers and industry decision-makers to make a judgment on the timing of integrated disease management practices. 
While laboratory-trained segmentation methods for disease symptom segmentation don't perform well in the field, they could still provide an automated disease method of laboratory experiments, lowering the requirements for highly skilled technicians with years of pathology experience. 
\\


\section*{Code availability}
The codes for the baseline reproduction are presented in \href{https://github.com/tqwei05/PlantSeg}{https://github.com/tqwei05/PlantSeg}. The codes benefit from \href{https://github.com/open-mmlab/mmsegmentation}{https://github.com/open-mmlab/mmsegmentation}, which provides a benchmark toolbox for numerous segmentation methods. 

\bibliography{sample}



\section*{Author contributions statement}

Tianqi Wei designed the study, built the dataset, conducted experiments and wrote the manuscript. Zhi Chen designed the study, built the dataset and wrote the manuscript. Xin Yu designed the study and revised the manuscript. Scott Chapman and Paul Melloy supervised the annotation process, validated the data and reviewed the manuscript. Zi Huang administrated the project, offered resources and reviewed the manuscript.

\section*{Competing interests} 

The authors declare no competing interests.






\end{document}